\documentclass[10pt,twocolumn,letterpaper]{article}

\usepackage[pagenumbers]{iccv} 

\definecolor{aliceblue}{rgb}{0.94, 0.97, 1.0}

\definecolor{iccvblue}{rgb}{0.21,0.49,0.74}
\usepackage[pagebackref,breaklinks,colorlinks,allcolors=iccvblue]{hyperref}
\definecolor{lightgray}{RGB}{220, 220, 220}
\usepackage{adjustbox}
\usepackage{multirow}
\usepackage{colortbl}
\usepackage{pifont}
\usepackage{xcolor}

\newcommand{\multiline}[3]{\begin{tabular}{@{}{#1}@{}}{#2}\\{#3}\end{tabular}}
\newcommand{\gray}[1]{\textcolor{gray}{#1}}

\def\paperID{6769} 
\def\confName{ICCV}
\def\confYear{2025}

\title{\includegraphics[height=1.4em]{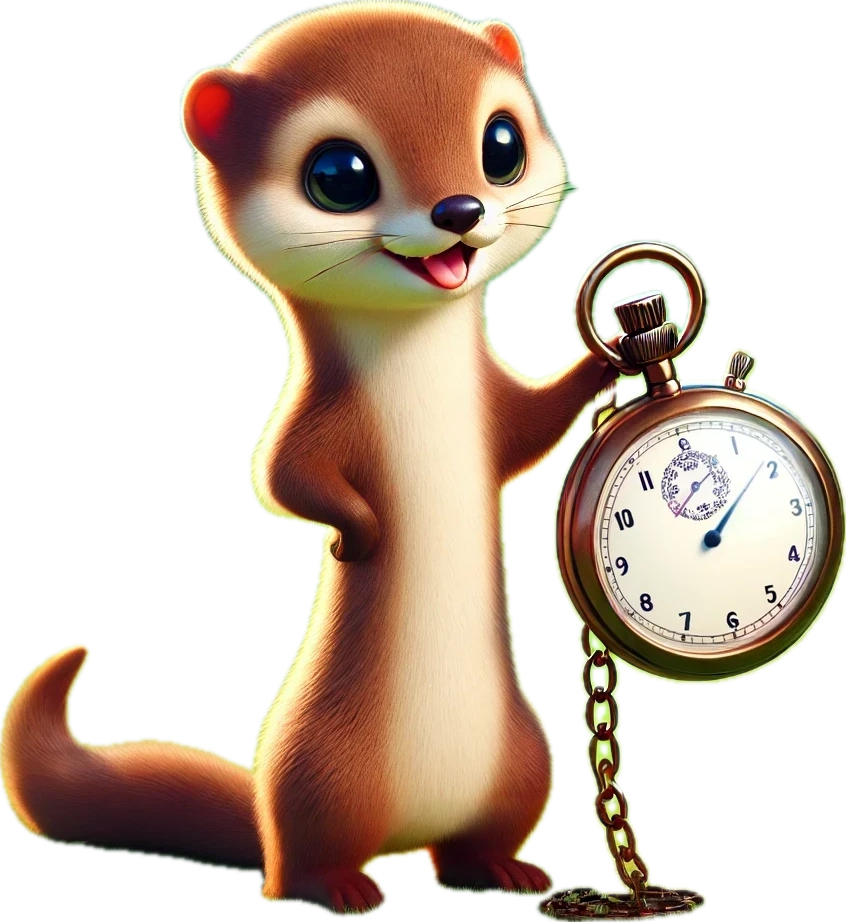}\hspace{0.2em}ST-VLM: Kinematic Instruction Tuning for \\ Spatio-Temporal Reasoning in Vision-Language Models}

\author{
Dohwan Ko\textsuperscript{\rm 1}$^{\dagger *}$\hspace{0.4cm}
Sihyeon Kim\textsuperscript{\rm 1}$^{*}$\hspace{0.4cm}
Yumin Suh\textsuperscript{\rm 2}$^{\ddag}$\hspace{0.4cm}
Vijay Kumar B.G\textsuperscript{\rm 2}\vspace{0.0cm} \\
Minseo Yoon\textsuperscript{\rm 1}\hspace{0.4cm} 
Manmohan Chandraker\textsuperscript{\rm 2,3}\hspace{0.4cm}
Hyunwoo J. Kim\textsuperscript{\rm 4}$^{\S}$\vspace{0.3cm} \\
\textsuperscript{\rm 1}Korea University\hspace{0.4cm}
\textsuperscript{\rm 2}NEC Labs America\hspace{0.4cm} 
\textsuperscript{\rm 3} UC San Diego\vspace{0cm}\hspace{0.4cm}
\textsuperscript{\rm 4}KAIST \\
\tt\small \{ikodoh, sh\_bs15, cooki0615\}@korea.ac.kr\vspace{0.0cm} \\
\tt\small \{yumin, vijay.kumar, manu\}@nec-labs.com\hspace{0.4cm} hyunwoojkim@kaist.ac.kr\vspace{0cm}
}

\begin{document}
\twocolumn[{
\maketitle
\begin{center}
    \centering
    \captionsetup{type=figure}
    \begin{subfigure}[b]{0.49\textwidth}
        \includegraphics[width=\textwidth]{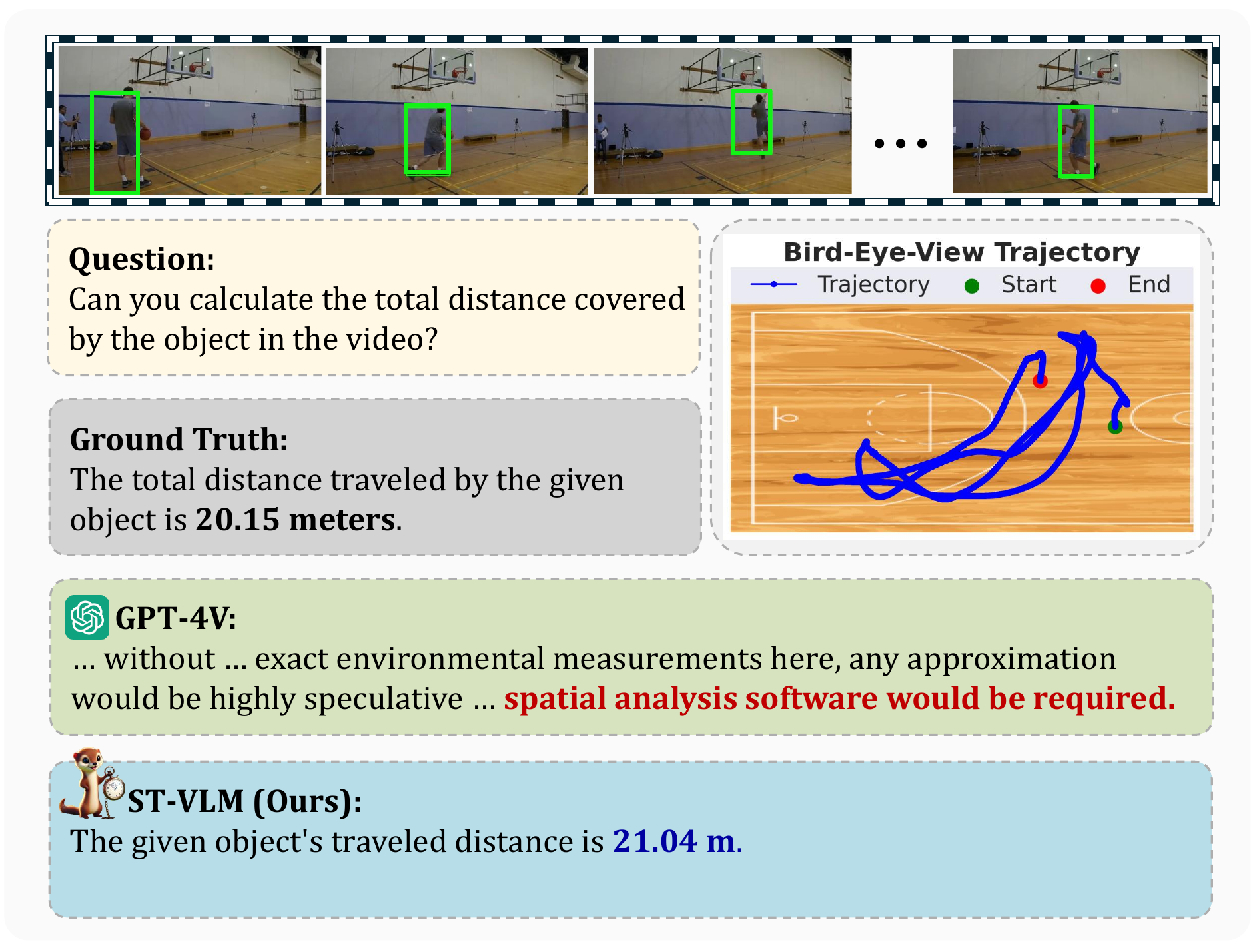}
        \caption{}
        \label{fig:intro2}
    \end{subfigure}
    \begin{subfigure}[b]{0.49\textwidth}
        \includegraphics[width=\textwidth]{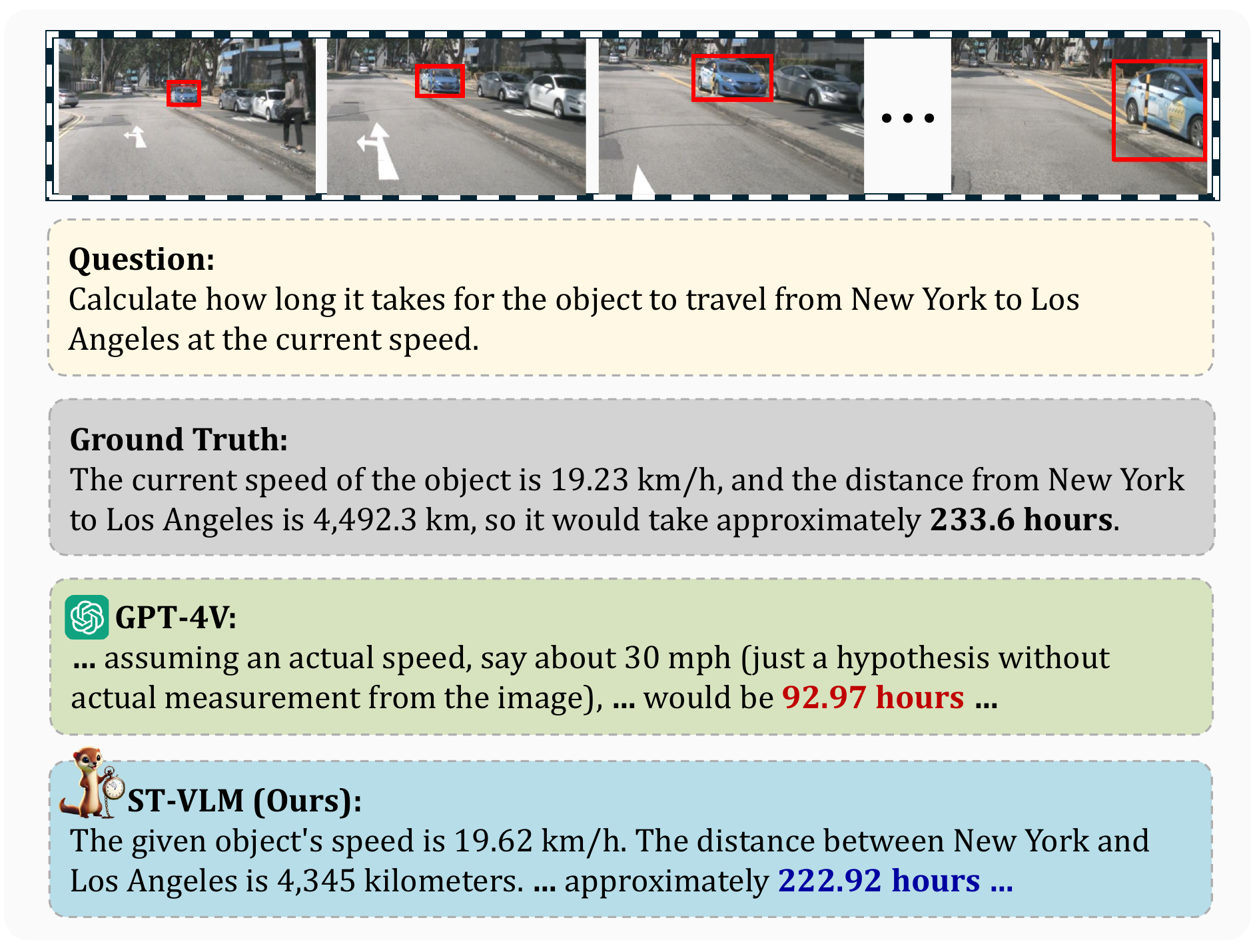}
        \caption{}
        \label{fig:intro3}
    \end{subfigure}
    \caption{
    Spatio-temporal reasoning in dynamic videos of moving objects.
    (a) \textbf{A challenging case with a complex trajectory}, asking the model to predict the total traveled distance using \textit{only} the video, without additional modalities such as 3D point clouds, depth map, or camera poses. 
    The video features a basketball player moving erratically across the court, making it more difficult for the model to predict. 
    (b) \textbf{An emerging capability of ST-VLM}, asking the model to solve multi-step reasoning questions that require integrating spatio-temporal understanding with its existing abilities (\eg, commonsense knowledge, logical reasoning, arithmetic computation).
    Since GPT-4V lacks spatio-temporal reasoning ability, it fails to generate accurate answers. 
    In (a), it responds that spatial analysis software is required, while in (b), it assumes a hypothetical speed of 30 mph, leading to a highly inaccurate answer compared to the ground truth. 
    In contrast, ST-VLM, equipped with spatio-temporal reasoning with the proposed STKit dataset, consistently provides accurate answers in both cases.
    }
    \label{fig:intro}
\end{center}
}]

\renewcommand{\thefootnote}{\fnsymbol{footnote}}
\footnotetext[0]{$\dagger$ Work partly done during an internship at NEC Labs America.} 
\footnotetext[0]{$*$ Equal contribution. $\ddag$ Currently at Atmanity Inc.}
\footnotetext[0]{$\S$ Corresponding author.}

\begin{figure*}[t]
    \centering
    \captionsetup{type=figure}
        \centering
        \includegraphics[width=0.98\textwidth]{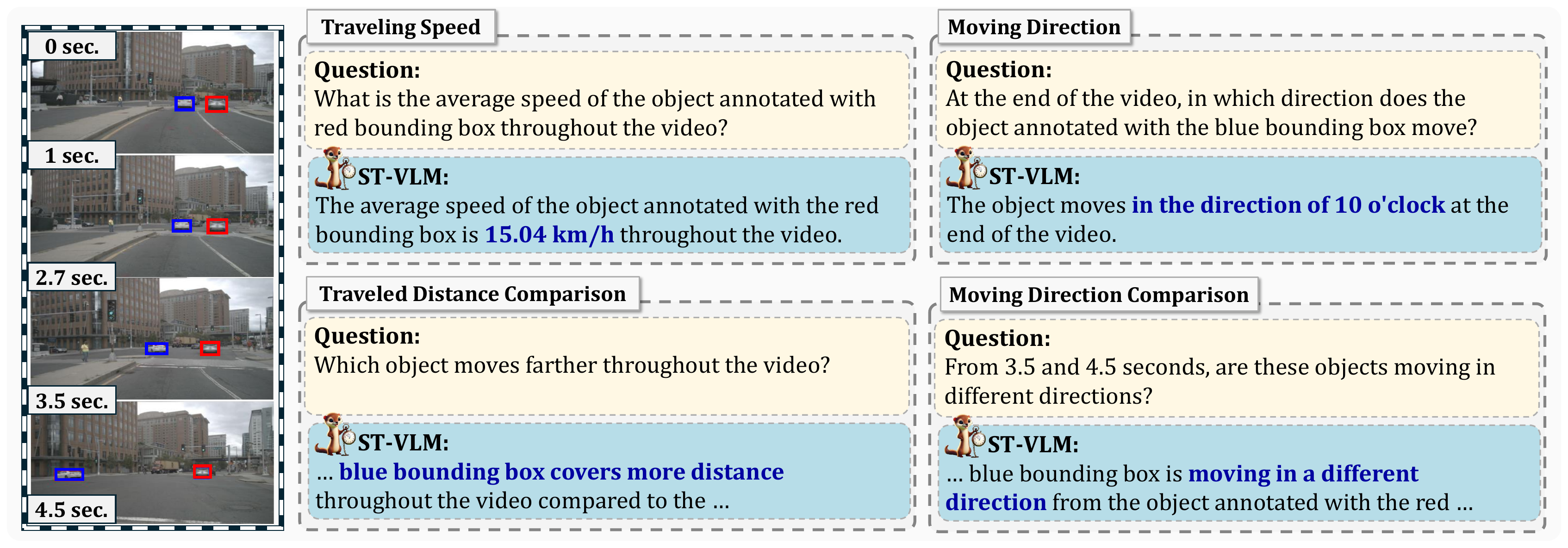}
        \vspace{-2.5mm}
        \caption{\textbf{Task examples from the proposed STKit-Bench along with predictions from ST-VLM}.
        }
        \vspace{-4.5mm}
    \label{fig:intro1}
\end{figure*}
\begin{abstract}
    Spatio-temporal reasoning is essential in understanding real-world environments in various fields, \eg, autonomous driving and sports analytics.
    Recent advances have improved the spatial reasoning ability of Vision-Language Models (VLMs) by introducing large-scale data, but these models still struggle to analyze kinematic elements like traveled distance and speed of moving objects.
    To bridge this gap, we construct a spatio-temporal reasoning dataset and benchmark involving kinematic instruction tuning, referred to as \textbf{STKit} and \textbf{STKit-Bench}.
    They consist of real-world videos with 3D annotations, detailing object motion dynamics: traveled distance, speed, movement direction, inter-object distance comparisons, and relative movement direction.
    To further scale such data construction to videos without 3D labels, we propose an automatic pipeline to generate pseudo-labels using 4D reconstruction in real-world scale.
    With our kinematic instruction tuning data for spatio-temporal reasoning, we present \textbf{ST-VLM}, a VLM enhanced for spatio-temporal reasoning, which exhibits outstanding performance on STKit-Bench.
    Furthermore, we show that ST-VLM generalizes robustly across diverse domains and tasks, outperforming baselines on other spatio-temporal benchmarks (\eg, ActivityNet, TVQA+).
    Finally, by integrating learned spatio-temporal reasoning with existing abilities, ST-VLM enables complex multi-step reasoning.
    Project page: \url{https://ikodoh.github.io/ST-VLM}.
\end{abstract}

\section{Introduction}

Spatio-temporal reasoning is the ability to infer spatial and temporal relationships within dynamic environments. 
For example, when analyzing a video of two cars driving on the road, spatio-temporal reasoning allows us to predict which car is moving faster or to estimate the movement direction and speed of a specific vehicle.
Incorporating this ability into Vision-Language Models (VLMs) enables complex reasoning that integrates spatio-temporal reasoning with multi-step inference.
Fig.~\ref{fig:intro3} illustrates that solving such problems involves accessing relevant knowledge (distances between cities), estimating kinematics (speed), understanding logical relationships (time = distance/speed), and performing calculations to derive the answer.
This high-level reasoning capability is essential in various applications, including autonomous driving, sports analytics, augmented/virtual reality, and embodied AI.
However, advanced AI models find it challenging to perform spatio-temporal reasoning; for instance, GPT-4V produces incorrect answers when asked to estimate the moving distance of a basketball player in the short video, as shown in Fig.~\ref{fig:intro2}.

Recent studies~\cite{chen2024spatialvlm,zhang2024countercurate,cheng2024spatialrgpt} have explored ways to improve the spatial reasoning capabilities of image-based VLMs, through large-scale data curation pipelines that annotate images with 3D spatial information, including object sizes and locations.
While effective for spatial reasoning, these approaches still fall short in spatio-temporal reasoning for video.
VLMs trained solely on spatial reasoning datasets struggle with tasks requiring temporal understanding, as they are limited to analyzing spatial relationships in images and cannot properly recognize temporal dynamics like kinematic quantities.
Building on the success of spatial reasoning in images through data curation pipelines~\cite{chen2024spatialvlm,cheng2024spatialrgpt}, we hypothesize that video datasets, especially those with dynamic object motion, annotated with kinematic spatio-temporal information is effective in enhancing the spatio-temporal reasoning abilities of VLMs.

In this paper, we propose a framework that goes beyond spatial reasoning in image-based VLMs to tackle spatio-temporal challenges in video-based VLMs. 
We first introduce STKit and STKit-Bench, spatio-temporal reasoning datasets and evaluation benchmarks designed for kinematic instruction tuning.
We define seven fundamental tasks in kinematic instructions, \eg, traveled distance and traveling speed (see Fig.~\ref{fig:intro1} for task examples).
We then present a pipeline for generating a kinematic instruction tuning dataset based on a spatio-temporal grounding framework in dynamic videos with 3D annotations, including driving videos~\cite{wilson2023argoverse, caesar2020nuscenes} with LiDAR-based point clouds and sports videos~\cite{grauman2024ego} with SLAM-based estimated point clouds from AR devices~\cite{engel2023project}.
Given the difficulty in acquiring point cloud-labeled videos, we also propose a pseudo-labeling pipeline based on 4D reconstruction for unlabeled videos~\cite{yu2020bdd100k,zhang2024video,li2021multisports} at a real-world scale.
By fine-tuning LLaVA-OneVision~\cite{li2024llava} on both labeled and pseudo-labeled kinematic instruction data, we develop ST-VLM which outperforms recent baseline VLMs on multiple spatio-temporal benchmarks.
Furthermore, equipped with spatio-temporal reasoning abilities based on kinematic understanding, our ST-VLM extends its capability to multi-step reasoning by integrating learned reasoning with VLM's existing abilities.


\begin{table*}[t!]
    \centering
    \begin{adjustbox}{width=\linewidth}
    \begin{tabular}{c|c|c|l}
        \toprule
        \textbf{Main Categories} & \textbf{Subcategories} & \textbf{Tasks} & \multicolumn{1}{c}{\textbf{Descriptions}} \\
        \midrule
        \midrule
        \multirow{8}{*}{\multiline{c}{Single}{Object}} & \multirow{4}{*}{Distance} & Traveled Distance & \multiline{l}{Predict the total traveled distance of the object given the timestamp.}{\gray{\eg, Can you calculate the total distance the object traveled between [START] and [END] seconds?}} \\
        \cmidrule{3-4}
        & & Traveling Speed & \multiline{l}{Predict the average traveling speed of the object given the timestamp.}{\gray{\eg, Tell me the object's average speed throughout the video.}} \\
        \cmidrule{2-4}
        & \multirow{4}{*}{Direction} & Movement Direction & \multiline{l}{Predict the movement direction of the object at the end of the video.}{\gray{\eg, What direction does the object travel at the end of the video?}} \\
        \cmidrule{3-4}
        & & Direction Timestamp & \multiline{l}{Predict the timestamp when the object moves in the given direction.}{\gray{\eg, Describe the timestamp when the object moves in the [DIRECTION] o'clock direction.}} \\
        \midrule
        \multirow{6}{*}{\multiline{c}{Multiple}{Objects}} & \multirow{4}{*}{Distance} & \multiline{c}{Traveled Distance}{Comparison} & \multiline{l}{Compare which object has traveled farther (or less).}{\gray{\eg, Which object travels a greater distance in the video?}} \\
        \cmidrule{3-4}
        & & \multiline{c}{Traveling Speed}{Comparison} & \multiline{l}{Compare which object has traveled faster (or slower).}{\gray{\eg, Which object moves faster throughout the video?}} \\
        \cmidrule{2-4}
        & Direction & \multiline{c}{Movement Direction}{Comparison} & \multiline{l}{Compare whether objects are moving in the same direction or not.}{\gray{\eg, Is object A moving in the same direction as object B in the video?}} \\
        \bottomrule
    \end{tabular}
    \end{adjustbox}
    \caption{\textbf{Overview of kinematic instructions.}
    For each task, a common prompt is prepended to provide additional information about the video timestamp and bounding box color of the object, \eg, ``The video lasts for $t$ seconds, and $n$ frames are uniformly sampled from it. These frames are located at $t_1, t_2, \dots, t_n$ seconds. There are $k$ objects annotated with [COLOR] bounding boxes in the video.''
    }
    \label{tab:category}
\end{table*}

\noindent In summary, our contributions are threefold:
\begin{itemize}
    \item[\textbullet] We introduce a new dataset and a benchmark, STKit and STKit-Bench, to endow VLMs kinematic understanding in dynamic videos, \eg, traveled distance and movement direction of the object, for spatio-temporal reasoning.
    \item[\textbullet] As video datasets with 3D annotations remain limited, we propose a pseudo-labeled data generation pipeline by leveraging 4D reconstruction on unlabeled videos.
    \item[\textbullet] We present ST-VLM, which significantly surpasses GPT-4V by 31.3\% on STKit-Bench with strong spatio-temporal reasoning. Our in-depth analyses demonstrate that ST-VLM excels in complex reasoning regarding object kinematics in various scenarios.
\end{itemize}


\section{Related work}

\noindent \textbf{Vision-Language Models (VLMs).}
Recent VLMs have demonstrated remarkable perception and reasoning capabilities across a wide range of tasks for both images~\cite{li2024llava,xu2024pllava,wang2024qwen2,lin2024vila} and videos~\cite{zhang2024llavanextvideo,wang2024internvideo2,cheng2024videollama,maaz2023video,li2023mvbench}, leveraging strong Large Language Models.
However, VLMs often fall short in understanding 3D geometry~\cite{liu2025mmbench}.
To overcome this issue, image-based spatial-aware VLMs~\cite{liu2025spatialcot,cai2024spatialbot,yang2024thinking,cai2024spatialbot,cheng2024spatialrgpt,chen2024spatialvlm} have been proposed to improve spatial reasoning capabilities.
For example, SpatialCoT~\cite{liu2025spatialcot} introduced chain-of-thought (CoT) spatial grounding, which exhibits advanced spatial reasoning.
For video-based VLMs, recent studies seek to enhance spatio-temporal capabilities~\cite{cheng2024videollama,li2025llava} for applications like autonomous driving~\cite{zhou2024embodied,wang2024omnidrive,ma2023dolphins} and embodied AI~\cite{huang2024rekep,cai2024spatialbot}.
In this paper, we propose a novel video-based VLM with high-level spatio-temporal reasoning capabilities regarding the object kinematics by directly estimating the traveled distance and movement direction.

\noindent \textbf{Spatio-temporal reasoning datasets.}
Several datasets have been proposed in the literature~\cite{lei2019tvqa+,zhang2020does} to improve the video-based VLMs' spatio-temporal reasoning ability.
TVQA+~\cite{lei2019tvqa+} is a video question answering dataset that requires localized spatial and temporal rationale to support the predicted answers, while VidSTG~\cite{zhang2020does} targets the spatio-temporal grounding given a query sentence. 
Several benchmarks have been introduced to evaluate video-based VLMs' spatio-temporal reasoning abilities in the general domain~\cite{li2023mvbench}, embodied AI~\cite{zhang2024mfe}, and autonomous driving~\cite{zhou2024embodied,wang2024omnidrive}.
However, these datasets and benchmarks do not explicitly take into account kinematics in dynamic videos.
In contrast, we focus on enhancing the spatio-temporal reasoning abilities of the video-based VLM with kinematics instruction tuning data.

\section{Method}

We aim to infuse VLMs with spatio-temporal reasoning capabilities through kinematic instruction tuning data called STKit.
First, in Sec.~\ref{subsec:kit}, we introduce seven tasks to categorize kinematic instructions in STKit.
We then present a spatio-temporal grounding framework to generate QA pairs in STKit with dynamic videos annotated with 3D point clouds in Sec.~\ref{subsec:lidar}.
Since 3D annotation is not always available, we propose a pseudo-labeling pipeline based on 4D reconstruction for unlabeled videos in Sec.~\ref{subsec:pseudo}.
Finally, we develop ST-VLM by fine-tuning LLaVA-OneVision~\cite{li2024llava} with STKit based on labeled and unlabled data in Sec.~\ref{subsec:training}.

\subsection{Kinematic instructions}
\label{subsec:kit}

We introduce STKit, a new kinematic instruction tuning dataset designed to enhance the spatio-temporal reasoning capabilities of VLMs.
Kinematic instructions include measuring kinematic quantities of objects in dynamic videos, such as their trajectories, traveled distances, and movement directions. 
We categorize kinematic instructions into seven tasks to comprehensively cover different aspects of spatio-temporal reasoning. 
These tasks are grouped into two main categories: Single Objects and Multiple Objects.
Each is further divided into two subcategories: Distance and Direction.
The seven spatio-temporal reasoning tasks are summarized in Tab.~\ref{tab:category}.
The tasks encourage the model to understand both the absolute distance and direction of an object's movement, as well as the relative distance and direction by comparing multiple objects.
Successfully managing these tasks requires the model to infer spatial information (\eg, object location) and temporal information (\eg, object movement), enabling complex spatio-temporal reasoning abilities built upon the prior knowledge of LLMs.

\begin{figure}[t] 
    \centering
    \includegraphics[width=\linewidth]{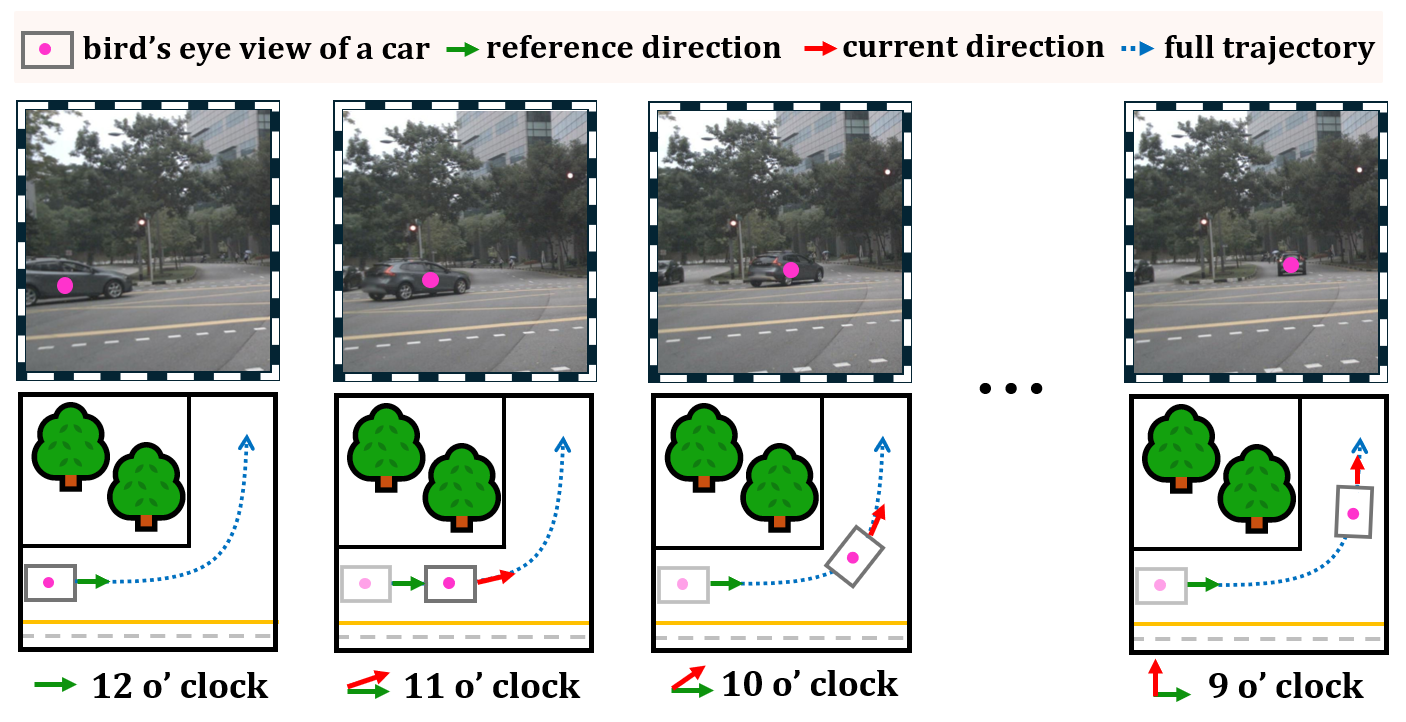}
    \vspace{-7.5mm}
    \caption{\textbf{Movement directions as clockwise directions.}
    }
    \vspace{-5.5mm}
    \label{fig:dir}
\end{figure}

\begin{figure*}[t!]
    \centering
    \includegraphics[width=\textwidth]{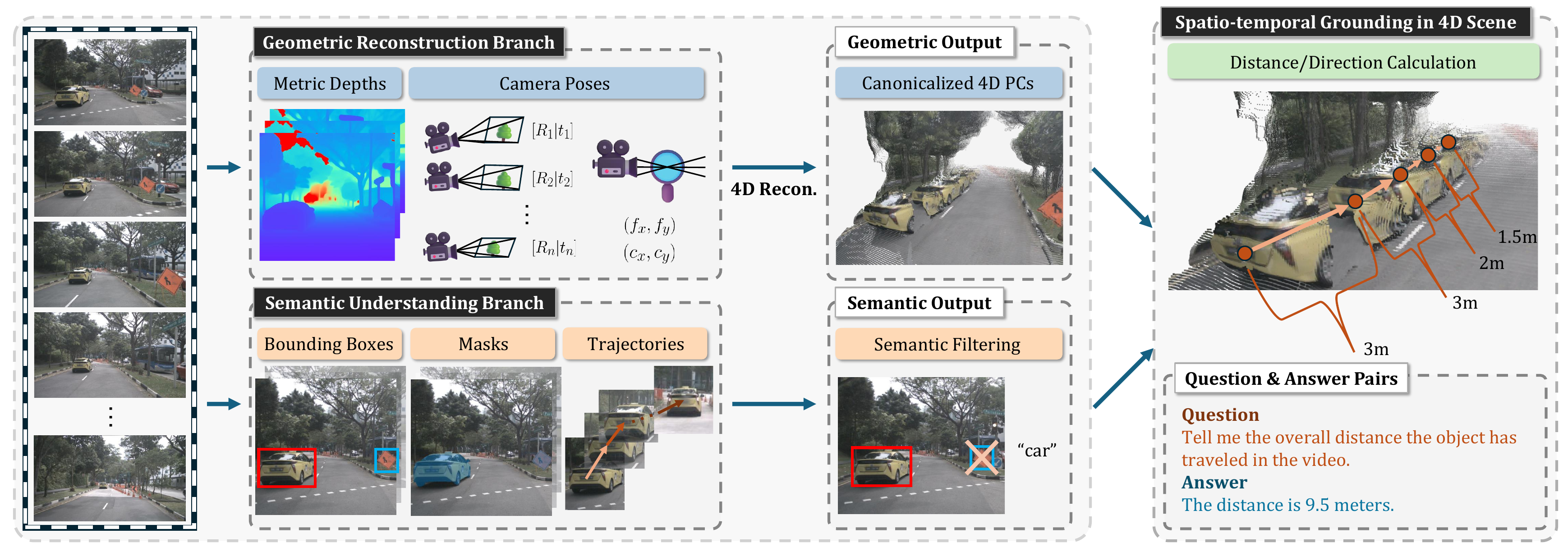}
    \caption{\textbf{Pseudo-label generation pipeline.}
    For the geometric reconstruction branch, a canonicalized 4D scene is reconstructed using MonST3R~\cite{zhang2024monst3r} and Metric3Dv2~\cite{hu2024metric3d}.
    For the semantic understanding branch, the object bounding boxes, segmentation masks, and trajectories are extracted using Grounded-SAM2~\cite{ren2024grounded}.
    Finally, by integrating each branch, 2D object masks are lifted to 3D and trajectories are computed by tracking 3D barycenters in the 4D scene.
    }
    \vspace{-5.5mm}
    \label{fig:main}
\end{figure*}

\subsection{Spatio-temporal grounding of dynamic videos}
\label{subsec:lidar}
\vspace{-1.5mm}
Generating QA pairs for STKit requires grounding the kinematic quantities of objects in dynamic videos. 
Videos with substantial object movement are most suitable for these tasks. 
Thus, we focus on diverse dynamic scenarios, such as autonomous driving and outdoor sports (\eg, football and basketball).
Specifically, we use autonomous driving datasets, such as Argoverse2~\cite{wilson2023argoverse} and NuScenes~\cite{caesar2020nuscenes}, which provide precise LiDAR-based 3D object coordinates represented in real-world scale at each timestamp. 
We also incorporate sports videos in Ego-Exo4D~\cite{grauman2024ego}, captured from wearable AR devices~\cite{engel2023project} that provide both RGB images and IMU, enabling accurate 3D trajectory estimation via Visual-Inertial Odometry (VIO) and SLAM. These estimates are treated as ground-truth in practice.

For each object annotated in the video, we have access to its 3D center and 3D bounding box coordinates in world space for each timestamp. 
Utilizing the 3D center coordinate $\mathbf{P}_t^{(i)}$ of $i$-th object at $t$ seconds, we first construct the trajectories by sampling the center at 0.5-second intervals over 40-frames videos, covering up to 20 seconds.
Then, we calculate the traveled distance of the $i$-th object between $s$ and $e$ seconds as the cumulative sum of distances between two consecutive frames, \ie, $\sum_{t=s}^{e-1} \| \mathbf{P}_{t + 1}^{(i)} - \mathbf{P}_t^{(i)}\|_2$.
The traveling speed is also calculated by dividing the total traveled distance by the duration $e - s$.

Calculating the movement direction for each object is more challenging than computing distance, as an absolute direction cannot be defined across all objects in the video. 
Hence, we establish a reference direction for each object based on its initial movement direction, calculated from the first two frames in which it appears, \ie, $\mathbf{P}^{(i)}_{s+1} - \mathbf{P}^{(i)}_{s}$.
Subsequent movement directions are computed as relative angles to this reference vector as:
\begin{equation}
    \theta_t = \arccos \left( \frac{(\mathbf{P}_{t+1}^{(i)} - \mathbf{P}_{t}^{(i)}) \cdot (\mathbf{P}_{s+1}^{(i)} - \mathbf{P}_{s}^{(i)})}{\|\mathbf{P}_{t+1}^{(i)} - \mathbf{P}_{t}^{(i)}\|\|\mathbf{P}_{s+1}^{(i)} - \mathbf{P}_{s}^{(i)}\|} \right).
\end{equation}

However, describing directions with angles is not intuitive, as humans typically do not use exact degrees. 
To make this more accessible for both humans and VLMs, we convert the calculated angles $\theta_t$ into discretized clockwise directions.
In detail, we set the initial reference direction as 12 o'clock, with subsequent directions expressed relative to this reference, as shown in Fig.~\ref{fig:dir}.
Due to the highly complex 3D trajectories in the sports domain, we exclude the movement direction category from that domain (see Fig.~\ref{fig:intro2} for a trajectory example).
\begin{table}[!t]
    \centering
    \label{tab:training_datasets}
    \begin{adjustbox}{width=\linewidth}
    \begin{tabular}{l|c c c c}
        \toprule
        \textbf{Dataset} & \textbf{\# QA pairs} & \textbf{\# Videos} & \textbf{Domain} & \textbf{3D annotation} \\
        \midrule
        \midrule
        NuScenes~\cite{caesar2020nuscenes} & 13K & 4K & autonomous driving & LiDAR\\
        Argoverse2~\cite{wilson2023argoverse} & 8K & 0.6K & autonomous driving & LiDAR\\
        Ego-Exo4D~\cite{grauman2024ego} & 6K & 0.8K & sports & VIO/SLAM\\
        BDD100K~\cite{yu2020bdd100k} & 86K & 15K & autonomous driving & - \\
        LLaVA-Video~\cite{zhang2024video} & 2K & 0.3K & general & - \\
        MultiSports~\cite{li2021multisports} & 1K & 0.2K & sports & - \\
        \bottomrule
    \end{tabular}
    \end{adjustbox}
    \caption{\textbf{Data composition of STKit.}
    We extract 21K dynamic videos and generate a total of 116K kinematic instructions from six datasets. 
    For videos without 3D annotations, we generate pseudo-labels via 4D reconstruction using Monst3R~\cite{zhang2024monst3r}.
    }
    \vspace{-6.5mm}
    \label{tab:compositions}
\end{table}

\vspace{-1.5mm}
\subsection{Pseudo-labeling for unlabeled dynamic videos}
\label{subsec:pseudo}
\vspace{-1.5mm}
Existing datasets with 3D annotations mostly cover self-driving scenarios, limiting applicability to broader domains. 
Thus, we propose a pseudo-labeling pipeline based on 4D reconstruction to extend STKit to different domains of unlabeled video datasets like BDD100K~\cite{yu2020bdd100k}, LLaVA-Video~\cite{zhang2024video}, and MultiSports~\cite{li2021multisports} as in Tab.~\ref{tab:compositions}. 
Leveraging recent advances in \textit{geometric reconstruction} and \textit{semantic understanding}, we reconstruct 4D scenes by lifting segmented objects from 2D frames into 3D space. 
By applying 4D reconstruction, we extend the spatio-temporal grounding in Sec.~\ref{subsec:lidar} to various domains, enabling accurate estimation of object kinematics.
See Fig.~\ref{fig:main} for the pipeline.

\noindent \textbf{Geometric reconstruction branch.}
For the 4D reconstruction given the unlabeled video, we utilize MonST3R~\cite{zhang2024monst3r} which proposes a 4D reconstruction framework that estimates scene geometry including depth and camera intrinsic/extrinsic, even in dynamic videos containing moving objects.
However, the reconstructed space by MonST3R is not aligned with the real-world scale, since it lacks a fixed reference for depth, resulting in reconstructions that are accurate in shape but arbitrary in size. 
This scale ambiguity poses a significant problem for our spatio-temporal reasoning tasks such as traveled distances and traveling speed.
To address the scale ambiguity, we integrate Metric3Dv2~\cite{hu2024metric3d} to obtain the absolute metric depth at the real-world scale.
Specifically, we canonicalize the reconstructed 4D scene by rescaling the original depth estimates from MonST3R to the metric depth from Metric3Dv2.

\noindent \textbf{Semantic understanding branch.}
We extract bounding boxes, segmentation masks, and trajectories of selected objects based on the open-vocabulary video semantic understanding model, Grounded-SAM2~\cite{ren2024grounded}.
We focus on classes of moving objects such as automobiles (\eg, cars, buses, trucks, motorcycles, bicycles) and humans.
To ensure the reliability of pseudo-labels, we filter detected objects based on confidence scores and bounding box sizes.

\noindent \textbf{Spatio-temporal grounding in canonicalized 4D scene.}
By integrating the outputs from the geometric reconstruction branch and the semantic understanding branch, the 2D segmentation mask of the selected objects is lifted into a 3D point cloud within the canonicalized 4D reconstructed scene.
We calculate the traveled distance, speed and movement direction for each object in the 3D space by tracking the barycenter of 3D object coordinates across video frames. 
Due to inherent limitations in monocular 4D reconstruction such as partial visibility and viewpoint constraints, we empirically develop filtering and smoothing strategies by comparing calculated trajectories against ground-truth trajectories in labeled datasets~\cite{caesar2020nuscenes}.
This validation allows us to establish effective criteria for filtering and smoothing, increasing the reliability of generated pseudo-labels for unlabeled datasets~\cite{yu2020bdd100k, zhang2024video, li2021multisports}.
See the supplement for details.
Hence, we scale up our spatio-temporal reasoning dataset with the pseudo-label generation pipeline.
\subsection{Kinematic instruction tuning}
\label{subsec:training}

Based on spatio-temporal groundings obtained by 27K point cloud labels (Sec.~\ref{subsec:lidar}) and 89K 4D reconstruction-based pseudo-labels (Sec.~\ref{subsec:pseudo}), we construct STKit for kinematic instruction tuning.
Specifically, with the distance and direction information, we adopt a template-based approach to construct QA pairs for our instruction-following dataset.
Tab.~\ref{tab:category} provides example templates for task categories.
To facilitate the model's understanding, we introduce a visual prompt by overlaying a bounding box on each frame, highlighting the object of interest (see Fig.~\ref{fig:intro}).
We also provide a common prompt including the timestamp and the color of the bounding box.
The generated QA pairs and the videos with bounding boxes are fed into the model for training. 

We develop ST-VLM by fine-tuning LLaVA-OneVision 7B~\cite{li2024llava}, which can accept various forms of visual inputs such as single images, multiple images, and videos, using our STKit dataset.
However, fine-tuning solely on STKit degrades performance on other generic benchmarks, implying overfitting to this specific task. 
Hence, we blend STKit with a subset of general supervised fine-tuning (SFT) datasets, LLaVA-Video-178K~\cite{zhang2024video}, which is a common practice in recent VLM training~\cite{lin2024vila,liu2024improved,cheng2024spatialrgpt}.
By blending these datasets, we empirically observe that our model can handle complex reasoning instructions that are not present in the predefined templates.
Detailed analyses are provided in Sec.~\ref{subsec:ifa}.
Furthermore, we adopt OpenSpatialDataset~\cite{cheng2024spatialrgpt} to enhance the model's spatial reasoning ability which is potentially advantageous in spatio-temporal reasoning.

\vspace{-3.5mm}
\section{STKit-Bench}
\vspace{-2.5mm}

\begin{table*}[!t]
    \centering
    \begin{adjustbox}{width=1.0\textwidth}
    \begin{tabular}{l|c c|c c|c c|c c|c|c|c|>{\columncolor{lightgray}}c}
        \toprule
        \multirow{5}{*}{\textbf{Models}} & \multicolumn{8}{c|}{Single Object} & \multicolumn{3}{c|}{Multiple Objects} & \multirow{4}{*}{\cellcolor{white} \textbf{Average}} \\
        \cmidrule{2-12}
        & \multicolumn{2}{c|}{\multiline{c}{Traveled}{Distance}} & \multicolumn{2}{c|}{\multiline{c}{Traveling}{Speed}} & \multicolumn{2}{c|}{\multiline{c}{Movement}{Direction}} & \multicolumn{2}{c|}{\multiline{c}{Direction}{Timestamp}} & \multiline{c}{Traveled Distance}{Comparison} & \multiline{c}{Traveling Speed}{Comparison} & \multiline{c}{Movement Direction}{Comparison} \\
        \cmidrule{2-13}
        & \textbf{Acc$\uparrow$} & \textbf{MAE$\downarrow$ (m)} & \textbf{Acc$\uparrow$} & \textbf{MAE$\downarrow$ (km/h)} & \textbf{Acc$\uparrow$} & \textbf{MAE$\downarrow$ (clock)} & \textbf{Acc$\uparrow$} & \textbf{IoU$\uparrow$} & \textbf{Acc$\uparrow$} & \textbf{Acc$\uparrow$} & \textbf{Acc$\uparrow$} & \cellcolor{white} \textbf{Acc$\uparrow$} \\
        \midrule
        \midrule
        \rowcolor[HTML]{FFF9C0}
        \multicolumn{13}{l}{\textbf{\textit{closed-source models}}} \\
        GPT-4o & 4.5 & 31.2 & 7.5 & 33.8 & 11.0 & 2.7 & 13.5 & 0.20 & 49.0 & 56.0 & 46.0 & 26.8 \\
        GPT-4V & 4.5 & 33.1 & 4.5 & 29.9 & 15.5 & 2.7 & 9.5 & 0.11 & 50.0 & 58.5 & 57.0 & 28.5 \\
        \midrule
        \rowcolor[HTML]{FFF9C0}
        \multicolumn{13}{l}{\textbf{\textit{autonomous driving-specific models}}} \\
        Dolphins~\cite{ma2023dolphins} & 16.0 & 241.9 & 2.5 & 32.0 & 5.0 & 3.0 & 1.5 & 0.07 & 37.0 & 19.0 & 41.0 & 17.4 \\
        ELM~\cite{zhou2024embodied} & 11.5 & 43.1 & 8.5 & 14.1 & 7.0 & 2.8 & 15.5 & 0.15 & 51.5 & 48.5 & 51.0 & 27.6 \\
        \midrule
        \rowcolor[HTML]{FFF9C0}
        \multicolumn{13}{l}{\textbf{\textit{open-source models}}} \\
        VideoLLaMA3-7B~\cite{zhang2025videollama} & 11.0 & 54.9 & 9.0 & 20.0 & 8.0 & 2.1 & 5.0 & 0.11 & 40.5 & 39.5 & 23.5 & 19.5 \\
        Qwen2.5-VL-7B~\cite{bai2025qwen2} & 4.5 & 206.4 & 7.0 & 26.1 & 3.0 & 2.8 & 2.0 & 0.04 & 47.5 & 47.5 & 46.0 & 22.5 \\
        InternVL2.5-8B~\cite{chen2024expanding} & 7.5 & 177.3 & 2.5 & 35.1 & 8.0 & 2.2 & 11.5 & 0.10 & 50.5 & 52.0 & 50.0 & 26.0 \\
        InternVideo2.5-8B~\cite{wang2025internvideo2} & 4.0 & 487.8 & 0.5 & 32.3 & 2.5 & 2.9 & 1.5 & 0.10 & 30.0 & 36.0 & 34.5 & 15.6 \\
        VideoChat-Flash-7B~\cite{li2024videochat} & 3.0 & 72.8 & 9.0 & 26.0 & 2.5 & 3.2 & 4.0 & 0.10 & 49.0 & 45.0 & 46.5 & 22.7 \\
        LLaVA-Video-Qwen2-7B~\cite{zhang2024video} & 7.5 & 74.0 & 10.5 & 24.2 & 13.5 & 2.3 & 2.0 & 0.05 & 50.5 & 45.5 & 44.5 & 24.9 \\
        LLaVA-OneVision-7B~\cite{li2024llava} & 8.0 & 60.2 & 7.5 & 23.9 & 4.5 & 2.0 & 23.5 & 0.22 & 43.0 & 52.5 & 53.0 & 27.4 \\
        \midrule
        \rowcolor{aliceblue}
        ST-VLM-7B (\textbf{Ours}) & \textbf{49.5} & \textbf{25.1} & \textbf{42.0} & \textbf{11.6} & \textbf{32.0} & \textbf{1.7} & \textbf{69.0} & \textbf{0.61} & \textbf{75.5} & \textbf{76.5} & \textbf{74.0} & \textbf{59.8} \\
        \bottomrule
    \end{tabular}
    \end{adjustbox}
    \caption{\textbf{Results on STKit-Bench.}
    The average accuracy is reported in the last column.
    }
    \vspace{-4.5mm}
    \label{tab:main}
\end{table*}
\begin{figure}[!t] 
    \centering
    \begin{subfigure}[h]{0.49\linewidth}
        \includegraphics[width=1.0\linewidth]{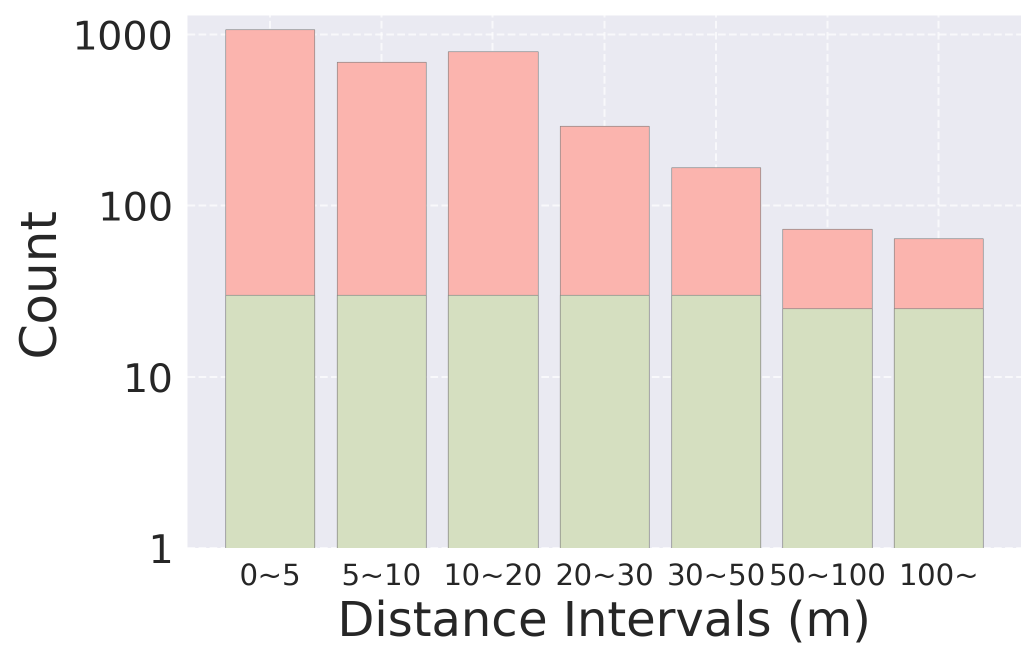}
        \caption*{}  
        \label{fig:statistics1}
    \end{subfigure}
    \begin{subfigure}[h]{0.49\linewidth}
        \includegraphics[width=1.0\linewidth]{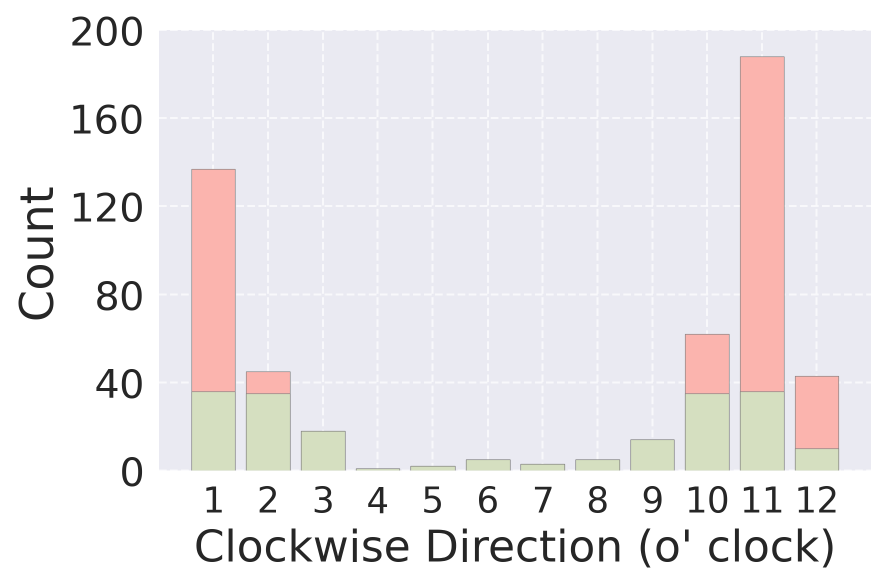}
        \caption*{}  
        \label{fig:statistics2}
    \end{subfigure}
    \vspace{-4.5mm}
    \caption{\textbf{Statistics of STKit-Bench.}
    We balance the number of samples for each label to prevent biased results.
    Red and green bars indicate the number of samples before/after balancing.
    }
    \vspace{-5.5mm}
    \label{fig:statistics}
\end{figure}

Since no benchmark exists for assessing the spatio-temporal reasoning capabilities of general VLMs, particularly in object kinematics, we introduce STKit-Bench, a benchmark comprising four datasets spanning the autonomous driving and sports domains. 
For autonomous driving, we incorporate NuPlan~\cite{caesar2021nuplan} for robust evaluation on an unseen dataset with NuScenes~\cite{caesar2020nuscenes} and Argoverse2~\cite{wilson2023argoverse}, which all provide LiDAR-based annotations. 
For sports, we use Ego-Exo4D~\cite{grauman2024ego} with SLAM-based point clouds.
We adopt the official validation splits to construct STKit-Bench, with dataset proportions of 74.8\% (NuPlan), 12.5\% (NuScenes), 5.6\% (Argoverse2), and 7.1\% (Ego-Exo4D).
Each task in STKit-Bench includes 200 QA pairs, resulting in a total of 1,400 QA pairs.
However, directly adopting generated QA pairs for the benchmark exhibits long-tail label distribution.
The red-colored bars in Fig.~\ref{fig:statistics} highlight the unbalanced label distribution in both distance and direction categories.
Therefore, to prevent biased evaluation results in STKit-Bench, we balance the number of samples for each label, as shown by the green-colored bars.

For evaluation, we use GPT-4 to extract the prediction from the response in natural language.
Then, we compare the prediction and the ground-truth answer and measure the performance by adopting the following metrics:

\noindent Given the ground-truth answer $y$ and the prediction $\hat{y}$,
(1) \textit{Traveled Distance} and (2) \textit{Traveling Speed}: Accuracy (correct if $y \times 0.75 \leq \hat{y} \leq y \times 1.25$) and MAE ($|y - \hat{y}|$).
(3) \textit{Movement Direction}: Accuracy (correct if $y =\hat{y}$ in the clockwise direction) and MAE ($\min(|y - \hat{y}|, 12 - |y - \hat{y}|)$ in the clockwise direction).
(4) \textit{Direction Timestamp}: Accuracy (correct if $\text{IoU}(y, \hat{y}) \geq 0.5$) and IoU.
(5) \textit{Traveled Distance Comparison}, (6) \textit{Traveling Speed Comparison}, and (7) \textit{Movement Direction Comparison}: Accuracy.

\noindent \textbf{Comparison with other benchmarks.}
Recently, several video benchmarks have been proposed in the literature~\cite{li2023mvbench,patraucean2023perception,fu2024video,zhou2024mlvu}.
For example, MLVU~\cite{zhou2024mlvu} aims to assess video-based VLMs for long-form video understanding, and VideoMME~\cite{fu2024video} focuses on the comprehensive perception ability of the model on a wide range of domains.
To tackle the problem that most VLMs overlook the temporal information, MVBench~\cite{li2023mvbench} has been proposed by covering diverse temporal understanding tasks, \eg, action sequence understanding, action prediction, and counterfactual inference.
More recently, several works~\cite{wang2024omnidrive,zhou2024embodied,ding2024holistic,nie2024reason2drive} have been introduced as a video spatio-temporal understanding benchmark for autonomous driving scenes.
In contrast, our STKit-Bench covers general scenes (\eg, sports) not limited to autonomous driving scenarios.
\section{Experiments}

In this section, we compare ST-VLM with baselines on STKit-Bench across diverse experimental settings to ensure a robust evaluation.

\subsection{Experimental settings}

\noindent \textbf{Baselines.}
To validate the spatio-temporal reasoning ability of ST-VLM, compared to strong proprietary models, 
we choose GPT-4V and GPT-4o as baselines.
Various Open-source video-based VLMs, including VideoLLaMA3~\cite{zhang2025videollama}, Qwen2.5-VL~\cite{bai2025qwen2}, InternVL2.5~\cite{chen2024expanding}, InternVideo2.5~\cite{wang2025internvideo2}, LLaVA-Video-Qwen2~\cite{zhang2024video}, and LLaVA-OneVision~\cite{li2024llava} are selected as baselines as well.

\noindent \textbf{Implementation details.}
For pseudo-labeling data, we construct 89K QA pairs from 15.5K dynamic videos sourced from BDD100K~\cite{yu2020bdd100k}, LLaVA-Video~\cite{zhang2024video}, and MultiSports~\cite{li2021multisports}. 
4D scene reconstruction with MonST3R requires approximately 400 seconds per video on a single A6000 GPU.
We train our model for 1 epoch with a batch size of 128.
The cosine learning rate scheduler is adopted with a learning rate of 1e-5 and the total training takes three days with 8 $\times$ A6000 GPUs.
For instruction tuning data composition, we blend STKit (117K) with subsets of 100K samples from LLaVA-Video-178K~\cite{zhang2024video} and 20K samples from OpenSpatialDataset~\cite{cheng2024spatialrgpt}, respectively.
\subsection{Results on STKit-Bench}

Tab.~\ref{tab:main} presents results on STKit-Bench, comparing ST-VLM to baseline VLMs.
Proprietary models, GPT-4V and GPT-4o, struggle with spatio-temporal reasoning on STKit-Bench.
In the Traveled Distance (TD) category, GPT-4V achieves an accuracy of only 4.5\% with a mean absolute error (MAE) of 33.1, indicating an average gap of 33.1~m from the ground-truth distance. 
Open-source models also face challenges with spatio-temporal reasoning.
For instance, LLaVA-OneVision~\cite{li2024llava} demonstrates poor performance with an average accuracy of 27.4\%.
Furthermore, we evaluate the performance of baseline models in autonomous driving, including Dolphins~\cite{ma2023dolphins} and ELM~\cite{zhou2024embodied}, both of which are trained on road scene videos.
However, we observe that both Dolphins and ELM are specifically trained with instructions for autonomous driving scenarios, making them unable to generalize to new types of instructions, resulting in poor performance with average accuracies of only 17.4\% and 27.6\%, respectively.
Our ST-VLM surpasses all baselines across seven tasks by a significant margin, \eg, achieving a 32.4\% higher average accuracy than LLaVA-OneVision, introducing new spatio-temporal reasoning capabilities which were absent in previous models.
\begin{table}[!t]
    \centering
    \begin{adjustbox}{width=\linewidth}
    \setlength{\tabcolsep}{4.0pt}
    \scriptsize
    \begin{tabular}{l|c c|c c c c|c c c|>{\columncolor{lightgray}}c}
        \toprule
        \textbf{Models} & $N$\textbf{-shots} & \textbf{geo.} & \textbf{TD} & \textbf{TS} & \textbf{MD} & \textbf{DT} & \textbf{TDC} & \textbf{TSC} & \textbf{MDC} & \cellcolor{white} \textbf{Avg} \\
        \midrule
        \midrule
        GPT-4V & 0 & \ding{56} & 4.5 & 4.5 & 15.5 & 9.5 & 50.0 & 58.5 & 57.0 & 28.5 \\
        GPT-4V & 1 & \ding{56} & 4.0 & 3.5 & 14.0 & 8.5 & 52.5 & 55.5 & 55.5 & 27.6 \\
        GPT-4V & 3 & \ding{56} & 8.0 & 5.5 & 16.0 & 49.5 & 55.0 & 49.0 & 46.5 & 32.8 \\
        \midrule
        GPT-4V & 0 & \ding{52} & 5.0 & 4.0 & 9.5 & 9.0 & 51.5 & 58.5 & 52.5 & 27.1 \\
        GPT-4V & 1 & \ding{52} & 3.5 & 4.5 & 9.0 & 7.5 & 47.5 & 49.5 & 47.5 & 24.1 \\
        GPT-4V & 3 & \ding{52} & 6.5 & 6.0 & 16.0 & 52.5 & 42.5 & 49.5 & 44.0 & 31.0 \\
        \midrule
        \rowcolor{aliceblue}
        ST-VLM (\textbf{Ours}) & - & - & \textbf{49.5} & \textbf{42.0} & \textbf{32.0} & \textbf{69.0} & \textbf{75.5} & \textbf{76.5} & \textbf{74.0} & \textbf{59.8} \\
        \bottomrule
    \end{tabular}
    \end{adjustbox}
    \caption{\textbf{Comparison with improved baselines.}
    We provide $N$ few-shot examples and additional geometric contexts (denoted as \textbf{geo.}), \ie, camera extrinsics and depth maps, to GPT-4V and report the accuracy for each task and the average accuracy.
    }
    \label{tab:prior}
\end{table}
\begin{table}[!t]
    \centering
    \begin{adjustbox}{width=\linewidth}
    \setlength{\tabcolsep}{4.0pt}
    \scriptsize
    \begin{tabular}{l|c c c c|c c c|>{\columncolor{lightgray}}c}
        \toprule
        \textbf{Models} & \textbf{TD} & \textbf{TS} & \textbf{MD} & \textbf{DT} & \textbf{TDC} & \textbf{TSC} & \textbf{MDC} & \cellcolor{white} \textbf{Avg} \\
        \midrule
        \midrule
        GPT-4o &  6.5 & 8.5 & 9.0 & 11.0 & 46.5 & 56.0 & 51.5 & 27.0 \\
        GPT-4V &  5.0 & 7.5 & 12.0 & 12.5 & 50.5 & 55.5 & 48.0 & 27.3 \\
        LLaVA-OneVision &  6.5 & 5.0 & 4.5 & 16.0 & 55.5 & 58.0 & 44.5 & 27.1 \\
        \midrule
        \rowcolor{aliceblue}
        ST-VLM (\textbf{Ours}) & \textbf{46.0} & \textbf{41.5} & \textbf{33.5} & \textbf{67.5} & \textbf{76.0} & \textbf{72.5} & \textbf{70.5} & \textbf{58.2} \\
        \bottomrule
    \end{tabular}
    \end{adjustbox}
    \caption{\textbf{Results on paraphrased STKit-Bench.} 
    }
    \label{tab:paraphrase}
\end{table}

\vspace{-1.0mm}
\begin{table}[!t]
    \centering
    \begin{adjustbox}{width=1.0\linewidth}
    \scriptsize
    \begin{tabular}{c c||c c c c|c c c|>{\columncolor{lightgray}}c}
        \toprule
        \textbf{Label} & \textbf{Pseudo-label} & \textbf{TD} & \textbf{TS} & \textbf{MD} & \textbf{DT} & \textbf{TDC} & \textbf{TSC} & \textbf{MDC} & \cellcolor{white} \textbf{Avg} \\
        \midrule
        \midrule
        - & - & 10.0 & 15.0 & 4.5 & 20.5 & 43.5 & 53.0 & 48.5 & 27.9 \\
        \ding{52} & - & 28.0 & 31.0 & 33.5 & 65.5 & 65.0 & 64.5 & 70.0 & 51.1 \\
        \rowcolor{aliceblue}
        \ding{52} & \ding{52} & \textbf{49.5} & \textbf{42.0} & \textbf{32.0} & \textbf{69.0} & \textbf{75.5} & \textbf{76.5} & \textbf{74.0} & \textbf{59.8} \\
        \bottomrule
    \end{tabular}
    \end{adjustbox}
    \caption{\textbf{Ablation on data composition.}
    }
    \label{tab:pseudo}
\end{table}

\subsection{Robust evaluation across diverse settings}

To ensure robust and fair evaluation, we present two experimental settings. 
First, we augment the baselines with contextual support for spatio-temporal reasoning and compare them to our method.
Specifically, we provide GPT-4V with QA pairs in a few-shot manner, and additionally give geometric context as inputs (\eg, camera extrinsics as textual prompts and depth maps as extra image inputs) to match the intrinsic geometric knowledge implicitly learned by ST-VLM from 3D point clouds.
As shown in Tab.~\ref{tab:prior}, increasing the number of few-shot examples indeed improves performance; however, giving extra geometric contexts leads to performance degradation. 
Nevertheless, GPT-4V's performance still remains significantly lower than that of ST-VLM.
Second, we verify the robustness of ST-VLM to variations in questions on STKit-Bench by paraphrasing questions using GPT-4.
In Tab.~\ref{tab:paraphrase}, ST-VLM maintains robust performance across paraphrased questions, outperforming other methods. 
This indicates that ST-VLM's strong performance arises from its inherent kinematic understanding rather than overfitting to specific question templates.

\begin{table}[!t]
    \centering
    \setlength{\tabcolsep}{2.0pt}
    \begin{adjustbox}{width=\linewidth}
    \begin{tabular}{l|c|c|c|c|c|>{\columncolor{lightgray}}c}
        \toprule
        \multirow{2}{*}{\textbf{Models}} & \textbf{PerceptionTest} & \textbf{MVBench} & \textbf{VideoMME} & \textbf{MLVU} & \textbf{NExT-QA} & \cellcolor{white} \textbf{Avg.} \\
        & val & test & w/o \& w/ subtitle & m-avg & test & \cellcolor{white} acc \\
        \midrule
        \midrule
        VILA-40B~\cite{lin2024vila} & 54.0 & - & 60.1 \& 61.1 & - & 67.9 & 60.8 \\
        PLLaVA-34B~\cite{xu2024pllava} & - & 58.1 & - \& - & - & - & 58.1 \\
        LLaVA-N-Video-34B~\cite{zhang2024llavanextvideo} & 51.6 & - & 52.0 \& 54.9 & - & 70.2 & 57.2 \\
        LongVA-7B~\cite{zhang2024long} & - & - & 52.6 \& 54.3 & 56.3 & 68.2 & 77.1 \\
        IXC-2.5-7B~\cite{zhang2024internlm} & 34.4 & 69.1 & 55.8 \& 58.8 & 37.3 & 71.0 & 54.4 \\
        LLaVA-N-Video-32B~\cite{zhang2024llavanextvideo} & 59.4 & - & 60.2 \& 63.0 & 65.5 & 77.3 & 65.1 \\
        \midrule
        LLaVA-OneVision-7B~\cite{li2024llava} & 57.1 & 56.7 & 58.2 \& 61.5 & 64.7 & 79.4 & 62.9 \\
        \rowcolor{aliceblue}
        ST-VLM-7B (\textbf{Ours}) & \textbf{64.5} & \textbf{57.4} & \textbf{60.5} \& \textbf{63.3} & \textbf{66.9} & \textbf{80.8} & \textbf{65.6} \\
        \bottomrule
    \end{tabular}
    \end{adjustbox}
    \caption{\textbf{Results on comprehensive video benchmarks.}
    }
    \label{tab:comprehensive}
\end{table}
\begin{figure}[t] 
    \centering
    \includegraphics[width=\linewidth]{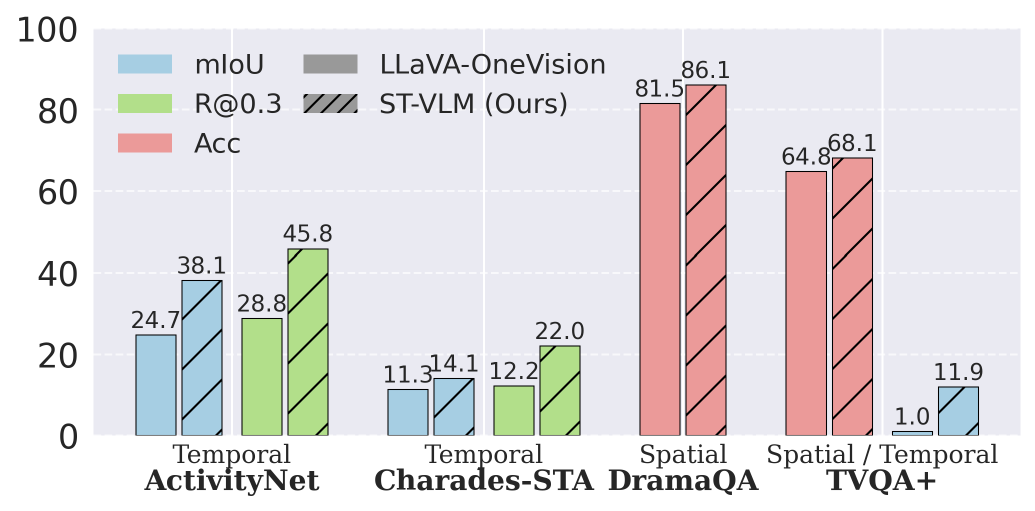}
    \caption{\textbf{Comparison on LLaVA-OneVision and ST-VLM for spatio-temporal understanding.}
    mIoU is multiplied by 100.
    }
    \vspace{-5.0mm}
    \label{fig:application}
\end{figure}
\begin{figure*}[t!] 
    \centering
    \begin{subfigure}[h]{0.49\linewidth}
        \includegraphics[width=\linewidth]{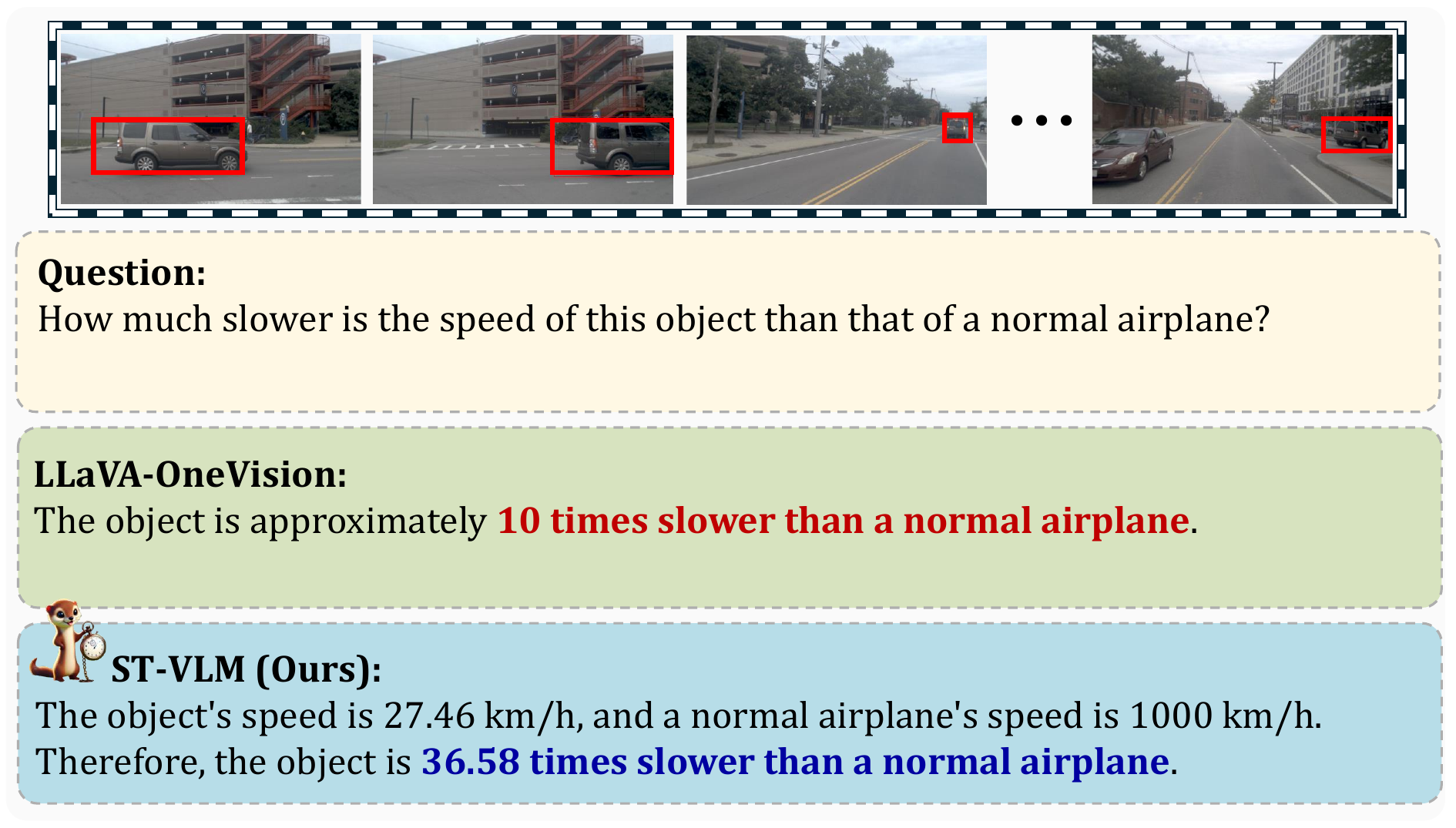}
        \caption{}
        \label{fig:qualitative1}
    \end{subfigure}
    \begin{subfigure}[h]{0.49\linewidth}
        \includegraphics[width=\linewidth]{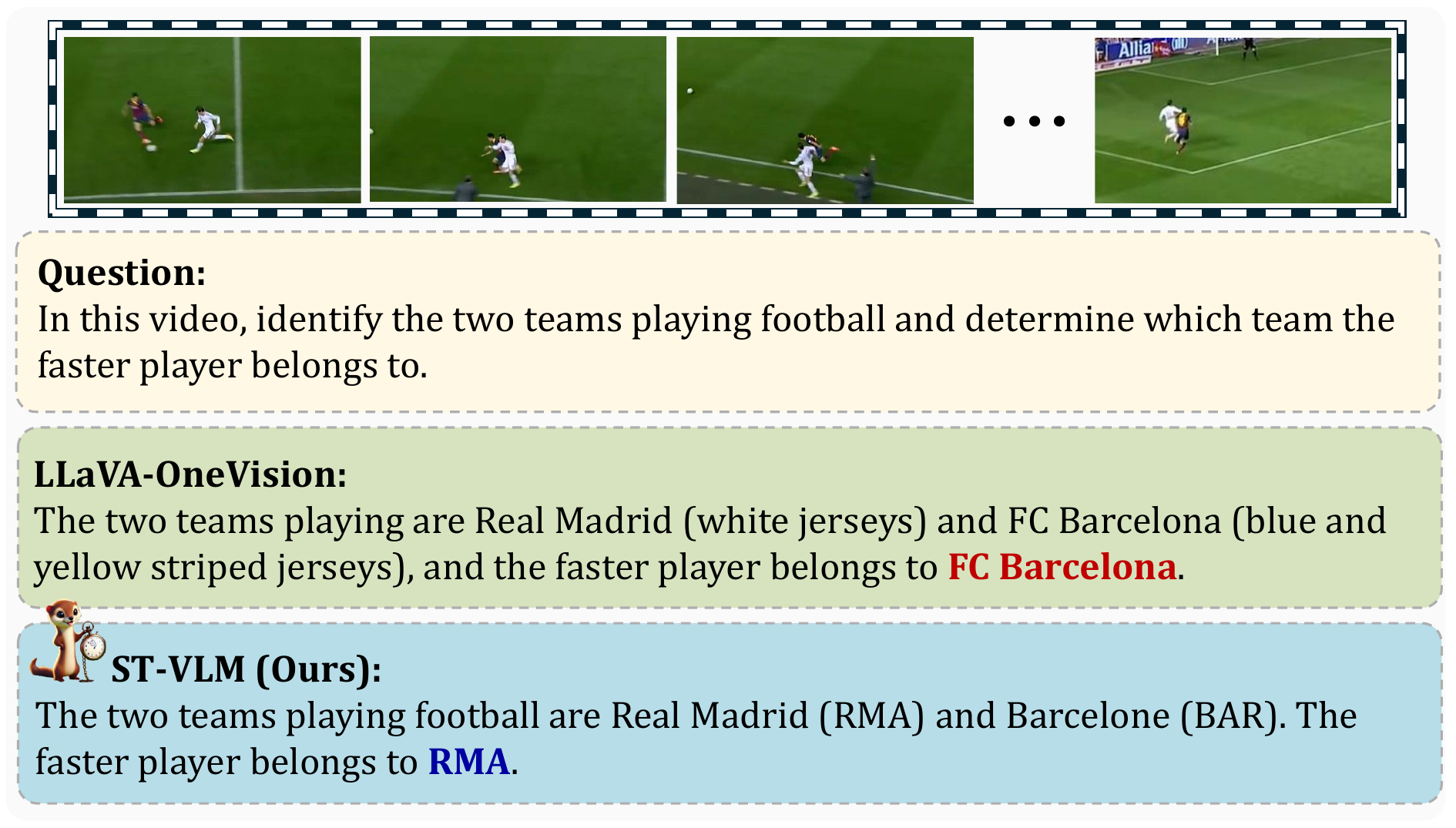}
        \caption{}
        \label{fig:qualitative2}
    \end{subfigure}
    \caption{\textbf{Qualitative results on emerging capabilities of ST-VLM with multi-step reasoning questions.}
    }
    \label{fig:qualitative}
\end{figure*}

\vspace{-2.0mm}
\section{Analysis}

In this section, we provide in-depth analyses to answer the following research questions: \\
\textbf{Q1.} How effective is our instruction tuning dataset, STKit? \\ 
\textbf{Q2.} Does the spatio-temporal reasoning ability of ST-VLM generalize across various domains and tasks? \\
\textbf{Q3.} Does ST-VLM exhibit emergent capabilities, combining spatio-temporal reasoning (\emph{learnt} ability) with LLM's knowledge (\emph{existing} ability) within multi-step reasoning?

\subsection{Data composition analysis of STKit}

We present the ablations on STKit introduced in Sec.~\ref{subsec:lidar} and \ref{subsec:pseudo} to discuss \textbf{Q1}.
In Tab.~\ref{tab:pseudo}, we compare the baseline performance of LLaVA-OneVision~\cite{li2024llava} against models fine-tuned with different data configurations.
Incorporating data made with labeled videos significantly improves the average performance from 27.9\% to 51.1\%. 
Furthermore, using pseudo-labeled data generated from unlabeled videos further enhances performance by 8.7\% improvement, highlighting the effectiveness of our 4D reconstruction-based pseudo-labeling pipeline.

\subsection{Generalized spatio-temporal understanding} 

We assess the generalization capacity of ST-VLM's spatio-temporal reasoning by evaluating it across various tasks and domains in a zero-shot setting for \textbf{Q2}.
First, Tab.~\ref{tab:comprehensive} shows that ST-VLM trained with STKit outperforms LLaVA-OneVision by 2.7\% in average accuracy even on general video understanding benchmarks, and it further surpasses LLaVA-N-Video-32B~\cite{zhang2024llavanextvideo} on average, which is the larger model than ours.
Notably, ST-VLM achieves a 7.4\% improvement on PerceptionTest~\cite{patraucean2023perception}, which is a benchmark that thoroughly evaluates the comprehension of fundamental physical properties in videos, by leveraging kinematic priors from STKit. 
See the supplement for further qualitative analyses.

Furthermore, Fig.~\ref{fig:application} compares the performance of our ST-VLM with LLaVA-OneVision on various spatio-temporal reasoning tasks other than proposed STKit-Bench.
For spatial reasoning, we design a region-aware VideoQA task by combining Referring Expression Generation and VideoQA. 
We reformulate existing VideoQA datasets, TVQA+~\cite{lei2019tvqa+} and DramaQA~\cite{choi2021dramaqa}, by providing a specific region and posing questions related to that region.
For temporal reasoning, we conduct temporal grounding tasks on TVQA+, ActivityNet~\cite{caba2015activitynet}, and Charades-STA~\cite{gao2017tall}, requiring the model to predict timestamps for query actions. 
Notably, TVQA+ involves both spatial and temporal reasoning, as it requires answers with supporting timestamps based on the question and specified region.
In Fig.~\ref{fig:application}, on temporal-aware tasks, ST-VLM demonstrates improved performance in both ActivityNet and Charades-STA datasets, measured by mean Intersection over Union (mIoU) and Recall at IoU 0.3 (R@0.3). 
For instance, R@0.3 on ActivityNet increases by 17.0\%. 
Additionally, in the spatial-aware task, ST-VLM outperforms LLaVA-OneVision in DramaQA by a margin of 4.6\% in accuracy. 
Finally, in TVQA+, both spatial and temporal grounding abilities of ST-VLM are improved compared to LLaVA-OneVision. 
\subsection{Emerging capabilities of ST-VLM}
\label{subsec:ifa}

Finally, we answer \textbf{Q3} through qualitative analyses in Fig.~\ref{fig:qualitative} and Fig.~\ref{fig:intro3}, showcasing ST-VLM's emerging multi-step reasoning capabilities involving spatio-temporal reasoning.
Despite not being explicitly trained for complex reasoning, 
ST-VLM effectively integrates kinematic reasoning with VLMs' existing abilities, such as knowledge retrieval, logical reasoning, and arithmetic computation.
For example, in Fig.~\ref{fig:qualitative1}, to answer, ``How much slower is this object's speed compared to a normal airplane?'', a model must (1) know the average speed of a normal airplane, (2) estimate the object's speed from the video, and (3) perform arithmetic to compare them. 
Leveraging kinematic reasoning, ST-VLM provides an accurate answer (36.58 times slower than a normal plane), while the LLaVA-OneVision offers less precise response (10 times slower than a normal plane).
Similarly, in Fig.~\ref{fig:qualitative2}, identifying the faster player requires recognizing teams by jerseys and estimating player speeds.
ST-VLM correctly does so, whereas the baseline fails.
These examples illustrate the effectiveness of our STKit-trained ST-VLM in multi-step reasoning that incorporates spatio-temporal understanding.

\vspace{-1.6mm}

\section{Conclusion}

\vspace{-1.8mm}

We present ST-VLM, a VLM with enhanced spatio-temporal reasoning capabilities, achieved through an enhanced understanding of kinematics in dynamic videos.
To this end, we introduce STKit and STKit-Bench, which define seven fundamental tasks using 3D-annotated video data.
Additionally, our 4D reconstruction-based data generation pipeline effectively mitigates the scarcity of 3D-annotated data, leading to improved performance on STKit-Bench.
Extensive analyses reveal that ST-VLM exhibits strong generalization across various video benchmarks and enables multi-step reasoning as an emerging capability, leveraging the combination of LLM's pretrained knowledge and newly acquired kinematic understanding.

{
    \small
    \bibliographystyle{ieeenat_fullname}
    \bibliography{main}
}

\end{document}